\documentclass[a4paper, 10pt, conference]{ieeeconf}      
\IEEEoverridecommandlockouts                              
\usepackage{packages}

\title{\LARGE\bf Feasibility Study of LIMMS,\\ A Multi-Agent Modular Robotic Delivery System with Various Locomotion and Manipulation Modes
}
\author{Taoyuanmin Zhu$^{\dagger 1}$, Gabriel I.~Fernandez$^{\dagger 1}$, Colin Togashi$^{1}$, Yeting Liu$^{1}$, and Dennis Hong$^{1}$
\thanks{
$^{\dagger}$Equal contribution by the first two authors for this work.\newline
\indent $^{1}$Robotics and Mechanisms Laboratory (RoMeLa), Department of Mechanical and Aerospace Engineering, University of California, Los Angeles, CA 90095, USA.
        {\tt\small \{tymzhu, gabriel808, ctogashi, liu1995, dennishong\}@ucla.edu}}
}

\begin{document}
\maketitle
\thispagestyle{empty}
\pagestyle{empty}

\begin{abstract}

The logistics of transporting a package from a storage facility to the consumer's front door usually employs highly specialized robots often times splitting sub-tasks up to different systems, e.g., manipulator arms to sort and wheeled vehicles to deliver. 
More recent endeavors attempt to have a unified approach with legged and humanoid robots.
These solutions, however, occupy large amounts of space thus reducing the number of packages that can fit into a delivery vehicle. 
As a result, these bulky robotic systems often reduce the potential for scalability and task parallelization.
In this paper, we introduce LIMMS (Latching Intelligent Modular Mobility System) to address both the manipulation and delivery portion of a typical last-mile delivery while maintaining a minimal spatial footprint. 
LIMMS is a symmetrically designed, 6 degree of freedom (DoF) appendage-like robot with wheels and latching mechanisms at both ends.
By latching onto a surface and anchoring at one end, LIMMS can function as a traditional 6-DoF manipulator arm. 
On the other hand, multiple LIMMS can latch onto a single box and behave like a legged robotic system where the package is the body. 
During transit, LIMMS folds up compactly and takes up much less space compared to traditional robotic systems. 
A large group of LIMMS units can fit inside of a single delivery vehicle, opening the potential for new delivery optimization and hybrid planning methods never done before.
In this paper, the feasibility of LIMMS is studied and presented using a hardware prototype as well as simulation results for a range of sub-tasks in a typical last-mile delivery.

\end{abstract}


\section{Introduction}
\label{sec:introduction}

The last mile of a package delivery tends to be the most difficult due to dynamic, uncertain environments both inside and outside of the vehicle. A delivery truck can start completely organized and packed with boxes. As packages are individually delivered, however, the neat stacks can slide or fall over as the vehicle moves. Outside of the vehicle, urban terrain varies based on both location and building type, e.g., delivering to a house with a porch versus an apartment complex that needs to be entered.
The task is further complicated by the fact that unloading packages from a vehicle often specifies a different set of design criteria from dropping packages off at a desired location. As a result, combinations of robots with separate specialities are being explored, such as, manipulators on top of quadrupeds \cite{adachi1996control,CHAI2022100029} or manipulators cooperating with wheeled robots \cite{nagasaka2010whole,ramalepa2021review}. These systems, though, come at a steep cost. In addition, most of these types of robots are meant for general purpose use and are not specifically optimized for package delivery.
\begin{figure}[t!]
		\centering
		\includegraphics[width=1\linewidth]{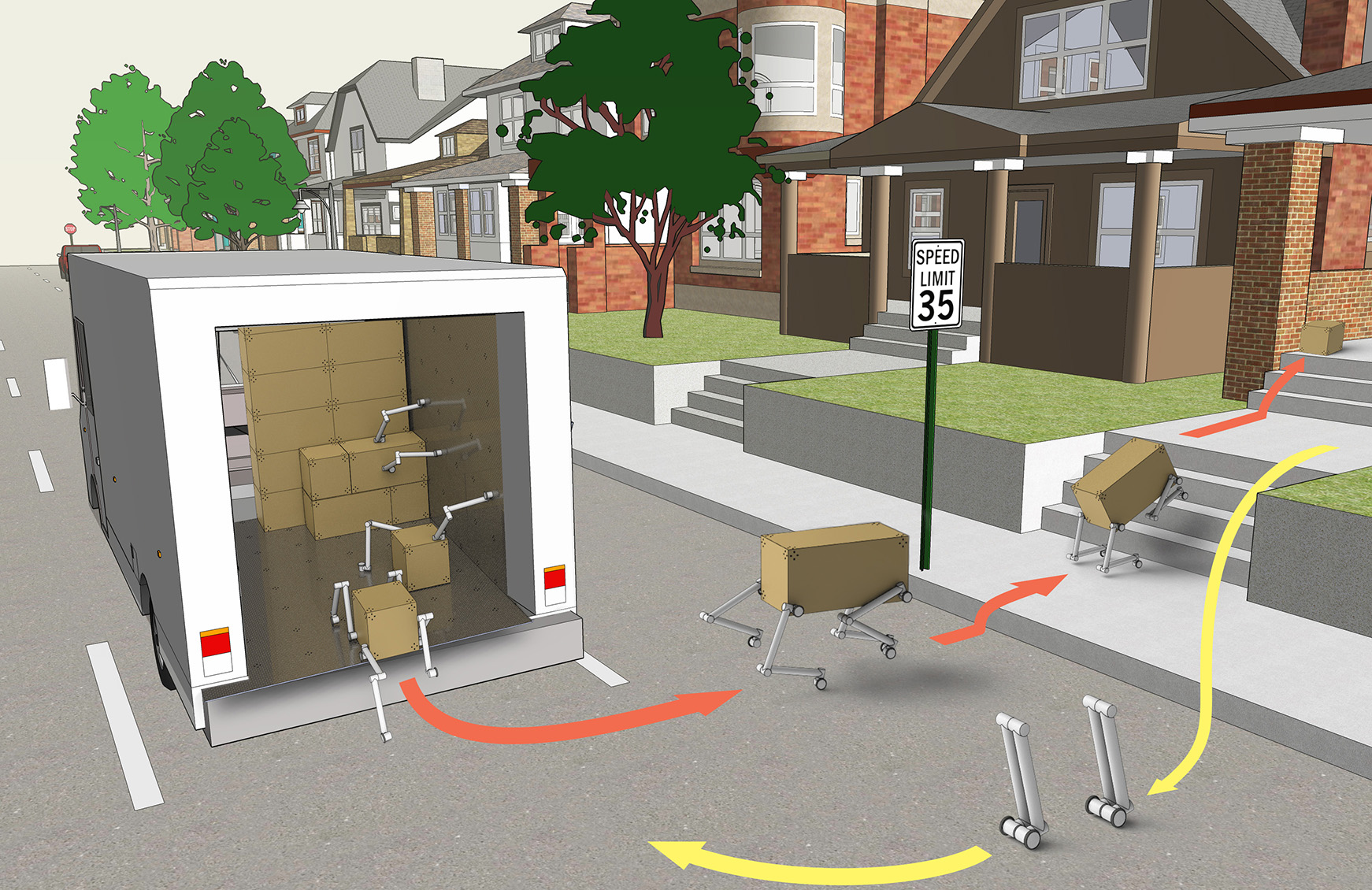}
		\caption {LIMMS concept depicted in a last-mile delivery task. Within the truck, LIMMS latches onto the wall and acts like a manipulator to move boxes to the exit of the truck. Multiple LIMMS then form a quadruped by using the delivery box as its body. After the package is successfully delivered, it returns as individual modules.}
		\label{fig:main}
\end{figure}

\begin{figure}[t]
    \centering
     \begin{subfigure}[]{0.49\linewidth}
         \centering
         \includegraphics[width=\textwidth]{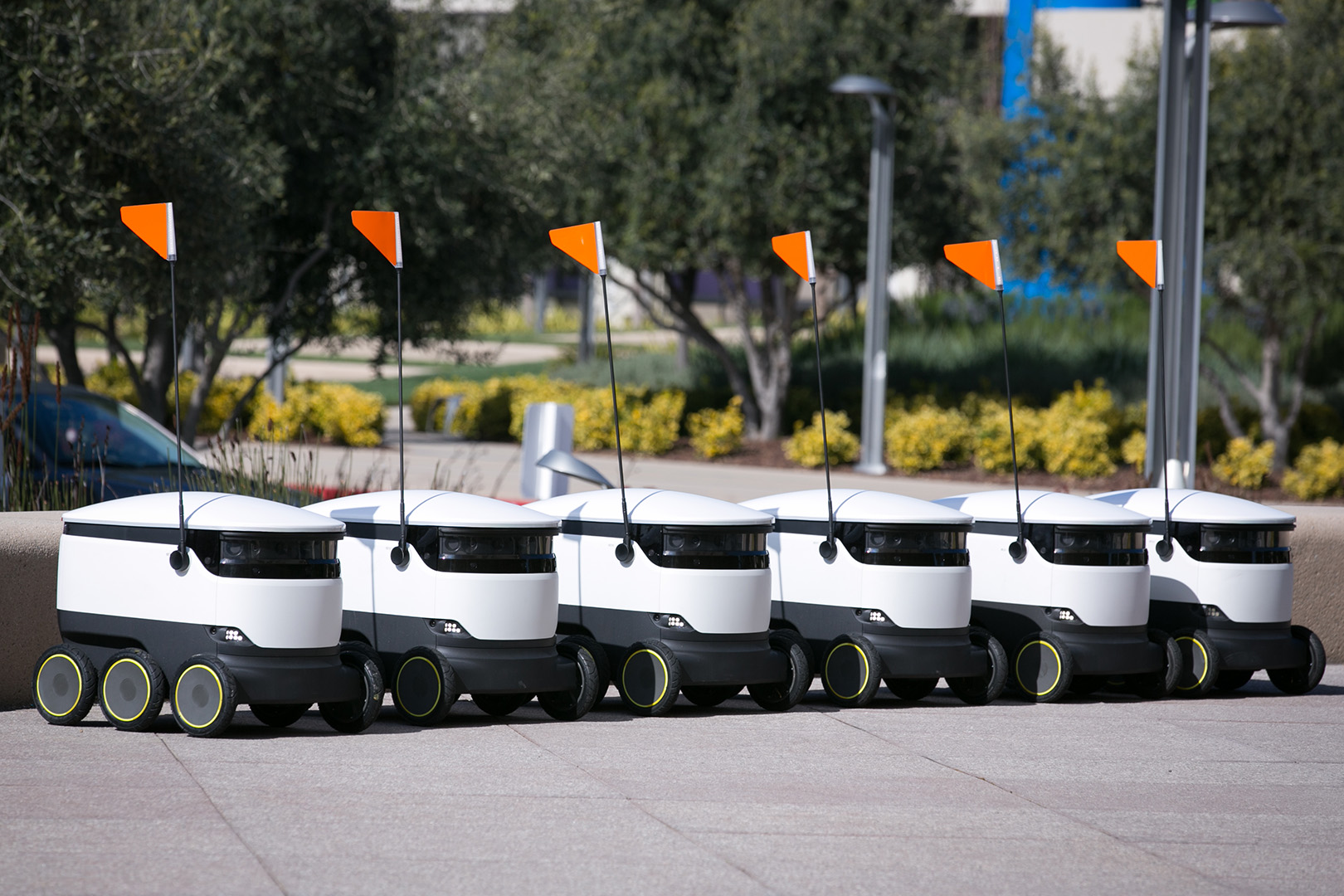}
         \caption{}
         \label{fig:starship}
     \end{subfigure}
     \hfill
     \begin{subfigure}[]{0.49\linewidth}
         \centering
         \includegraphics[width=\textwidth]{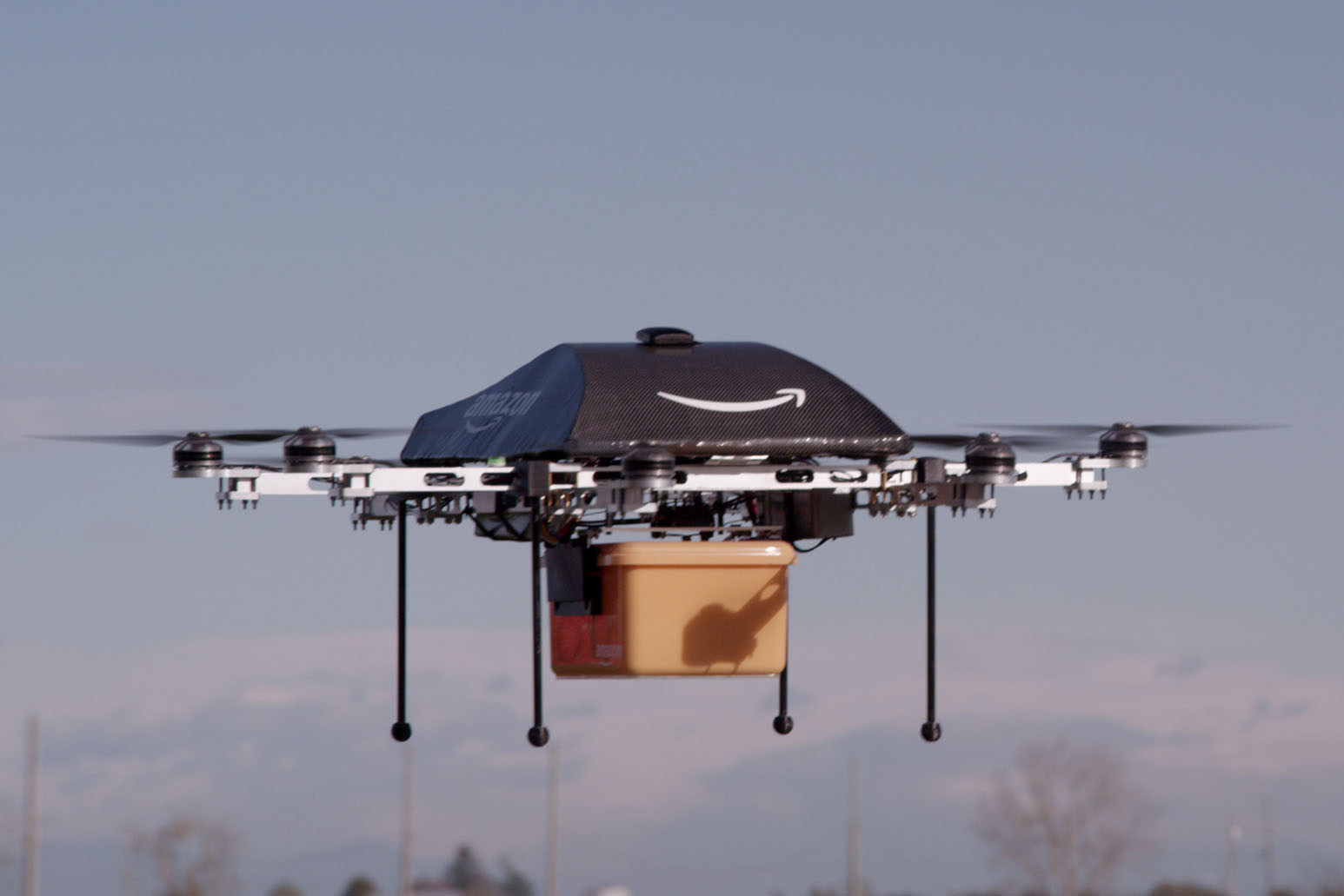}
         \caption{}
         \label{fig:primeair}
     \end{subfigure}

     \begin{subfigure}[]{0.49\linewidth}
         \centering
         \includegraphics[width=\textwidth]{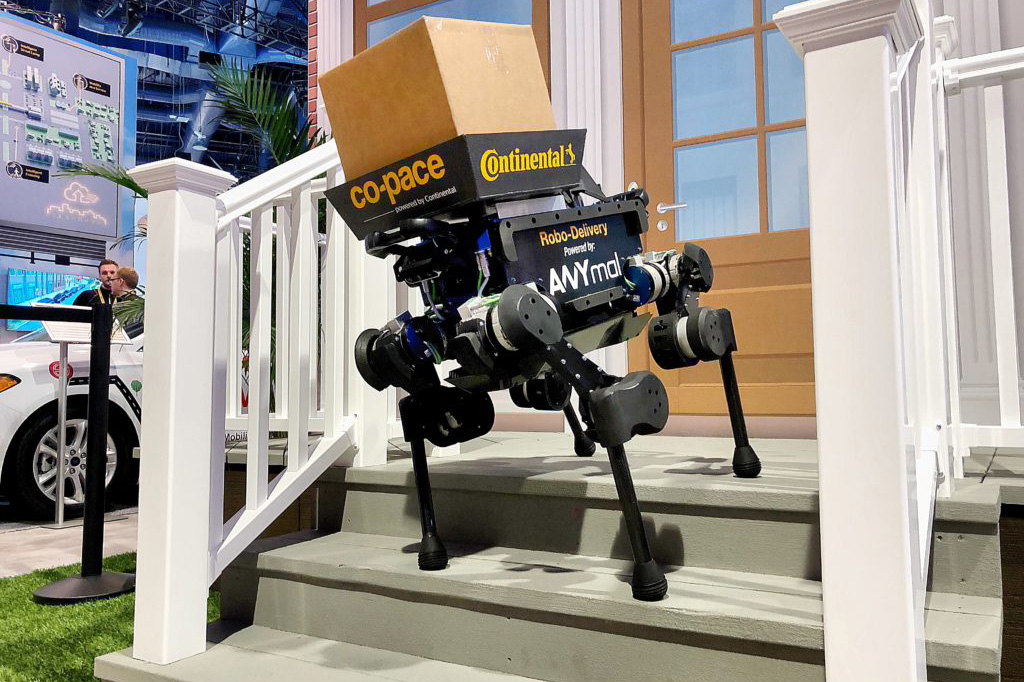}
         \caption{}
         \label{fig:anymal}
     \end{subfigure}
     \hfill
     \begin{subfigure}[]{0.49\linewidth}
         \centering
         \includegraphics[width=\textwidth]{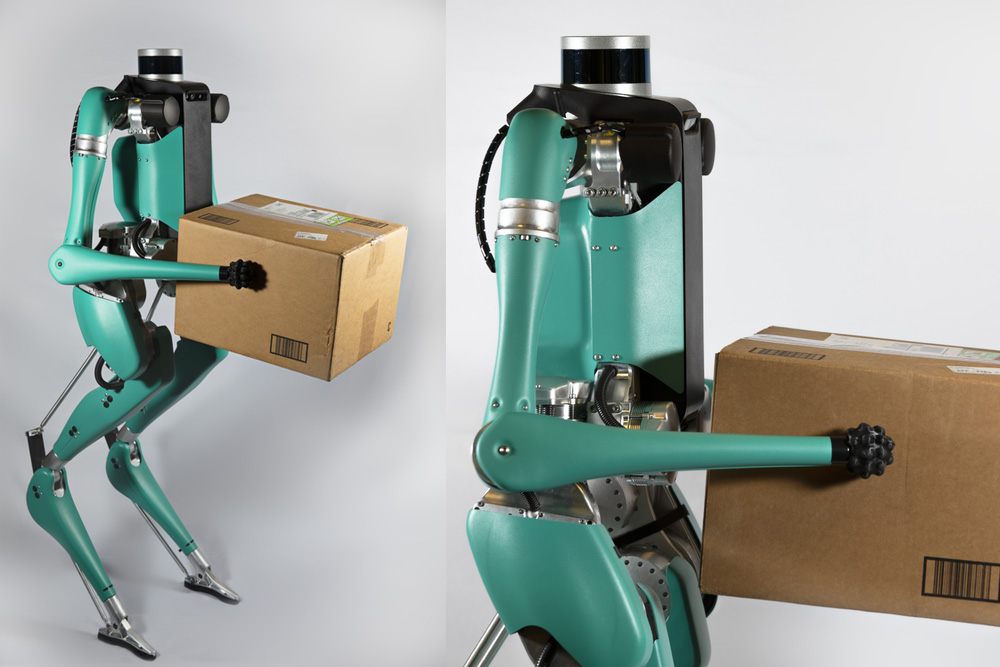}
         \caption{}
         \label{fig:digit}
     \end{subfigure}
     
    \caption{Existing delivery robotic solutions: (a) Starship delivery robot \cite{starship}, (b) Amazon Prime Air drone delivery \cite{amazon}, (c) ANYmal quadruped robot \cite{anymal}, (d) Digit bipedal robot \cite{digit}. }
    \label{fig:existing_robots}
 \end{figure}

Current practical solutions usually only solve a subset of the actual problem. For example, Starship robot \cite{starship}, as shown in \cref{fig:starship}, is a six wheeled delivery robot that requires couriers or businesses to place packages in its cargo bay and recipients to remove the package on arrival. The robot is only responsible for package transportation. Other wheeled robots handle even more specialized tasks, such as  Hikrobots \cite{hikrobot} which operate in warehouses and sort packages by moving the shelves they are stored in.

By nature wheeled robots are constrained to a single plane and flat terrain, though very efficient in those environments. In this sense, they often struggle with handling tasks with uncertain environments in delivery logistics. Moreover, the two dimensional planar constraint often leads to the need for complex coordination and planning to avoid bottlenecks. Effectively, wheels ground the delivery system reducing the overall operational space at any given moment for other robots to use. The smaller the available region, the more of a necessity proper coordination is to maintain efficient robot use. If too many robots are confined in a small space, multiple wheeled robots can become more inefficient than a single robot in what could amount to a large traffic jam.  Drones, on the other hand, fair much better in this sense since they can operate in all three dimensions. Services like Amazon Prime Air \cite{amazon}, see \cref{fig:primeair}, could take off if they were not limited by operational and safety regulations.

Meanwhile, robot systems that handle a wider scope of the delivery process are often much larger and more complex. For instance, quadrupeds, such as ANYmal in \cref{fig:anymal}, a four legged robot \cite{anymal,CHAI2022100029}, have the ability of traversing many different types of terrain that a wheeled robot cannot operate on. They also gain the ability to move in different ways, e.g., twist their body to unload a package on the ground. ANYmal has also been modified with wheels to provide the benefits of quicker locomotion over flat ground \cite{bjelonic2020whole}. A quadruped can be viewed as a more stable, less complex version of a humanoid robot, since two of its legs can be converted to arms \cite{hooks2020alphred}. This would be ideal since humanoid robots could theoretically traverse and accomplish similar tasks as humans. As shown in \cref{fig:digit}, humanoid robotic platforms such as Digit are being actively developed for last-mile delivery \cite{digit}.

A major drawback of legged systems is that they have a higher cost of transport \cite{keeprollin}, reducing the number of deliveries performed before a required recharge. On top of that, they also tend to occupy large amounts of space that could be devoted to more packages. This has a compounding effect of reducing the total capacity of packages per delivery an automated delivery vehicle can handle. Thus, for last-mile delivery, legged systems may not be cost efficient enough for adoption.

To mitigate these spatial and scaling limitations, we propose LIMMS (Latching Intelligent Modular Mobility System), an atypical modular robotic system for last-mile autonomous delivery. 
A single LIMMS unit is a 6-DoF arm-like robot with latching mechanisms and wheels at each end. The latches allow either side to anchor onto surfaces to become the base of the robot while the wheels can be used for locomotion. Each LIMMS unit acts as a robotic appendage and can take on a variety of roles by latching its base link to various surfaces. In this sense, multiple LIMMS can cooperate and take on different roles to achieve a goal in an efficient manner.

As far as the authors are aware, similar robotic arm concepts can only be found in spacecraft applications where its scale is more massive. Mobile Servicing System (MSS) and European Robotic Arm (ERA) from the International Space Station (ISS) both are equipped with latching mechanisms on either end of the arm \cite{stieber1999overview}\cite{cruijssen2014european}. Other latching robots, such as modular robots, focus on connecting between themselves to reconfigure to a need \cite{hasbulah2019comprehensive}, whereas LIMMS mainly interacts with its environment to increase leverage or manipulate objects.

A typical delivery routine is shown in \cref{fig:main}. The delivery truck would be lined with anchor points in a grid-like fashion along the walls, floor, and roof. Meanwhile, cardboard delivery boxes would be fabricated with similar mating patterns on each corner of the box. By securing a base at an anchor point and the other base to a box, one or multiple LIMMS could behave as manipulators to move boxes towards the exit of the truck. Meanwhile, four LIMMS could attach themselves to a delivery box and form a quadruped such that the box becomes the structure of the body. The quadruped would then deliver itself to the destination, avoiding any obstacles along the way. The wheels at the end of the \textit{legs} make it possible to have much faster and energy efficient locomotion on flat ground while maintaining the ability to walk over uneven terrain. After delivering the package and detaching from the box, LIMMS can either connect with each other for legged locomotion or move individually in its self-balancing wheel mode and return to the delivery truck for the next package.

In this paper, we introduce LIMMS as a concept for the logistics industry, as well as show the feasibility of such a system for last-mile delivery. Furthermore, LIMMS first hardware prototype is revealed. Finally, we demonstrate several modes of operation working in a simulated environment.
To summarize, our contributions are as follows:
\begin{enumerate}
    \item Introduce LIMMS, a novel modular robot concept,
    \item Detail its design with analysis and presented an early stage prototype of a LIMMS unit, and
    \item Verify the feasibility of its modes in simulation. 
\end{enumerate}

The rest of the paper is organized as follows: \cref{sec:requirements} explains functional requirements, 
\cref{sec:design} details the design and analysis of the LIMMS hardware prototype,
\cref{sec:simulation} verifies sub-tasks in a simulated environment for feasibility, and finally, \cref{sec:conclusion} concludes the paper and lays out potential future work.

\section{Requirements and Assumptions}
\label{sec:requirements}

For a typical last-mile delivery scenario, design requirements will be focused on the size and weight of the package, distance of travel for the robot as well as typical obstacles encountered by the robot.
According to public statistics, 86\% of Amazon’s packages are less than 5 lbs \cite{bamburry2015drones}. A target assumption of 2 kg lifting capacity will cover a significant portion of packages for a typical e-commerce company. It should be noted that due to the modular design of LIMMS, packages of much higher weight can still be handled by a system of multiple LIMMS working in tandem. 
 
For the first iteration of the LIMMS prototype, we considered boxes with a unit length of 12 in ($\approx$30 cm), i.e. box dimensions will always be in multiples of 12 in with the smallest box being 12 in $\times$ 12 in $\times$ 12 in. This assumption makes more efficient box stacking possible.
Additionally, we presume an average of 50 m of travel from the delivery vehicle to the destination and back.
The most common obstacles the robot will encounter are curbs and stairs. Standard curb height in the United States is 6 in ($\approx$15 cm) \cite{curb}. According to California Building Code, the maximum stair height is 7 in ($\approx$18 cm) with minimum depth of 11 in ($\approx$28 cm) \cite{stairs}. To assess the duty cycle of the robot, the following schedule is used:
\begin{enumerate}
    \item Idle in delivery vehicle and charging battery (15 min).
    \item Locate and attach to the delivery box (5 min).
    \item Enter quadruped mode and get perception data (3 min).
    \item Use wheels or trot towards destination (5 min).
    \item Detach and return back to the delivery vehicle (5 min).
\end{enumerate}

For the purposes of studying the feasibility of LIMMS, the focus has been put on mechanical design and locomotion control. Electronics, perception sensors, and computational power are considered outside the scope of this paper and will be considered in the future.

%


\section{Hardware design}
\label{sec:design}
Considering the complex nature of executing both manipulation and locomotion tasks, it is favorable to have as many DoFs as possible without overburdening the robot with weight and size. In total, LIMMS has 6 DoFs configured in a nontraditional symmetric fashion. 
Each end of LIMMS is equipped with a latching mechanism for manipulation as well as a wheel for locomotion. A single LIMMS weighs $4.14$ kg and can be contained within a $0.43$ m $\times$ $0.22$ m $\times$ $0.18$ m box when folded up as in \cref{fig:kinematics}. When fully stretched, it can reach up to $0.75$ m.

\subsection{Joint configuration}
\label{sec:joint_config}
A traditional 6-DoF robotic manipulator arm consists of a 2-DoF yaw-pitch base, an elbow joint, and a 3-DoF wrist. This joint configuration is preferred with the base and elbow motors providing the lifting power while the wrist corrects for the desired orientation. However, the base flipped in a similar fashion to how LIMMS can operate, a traditional arm would now have a 3-DoF base and 2-DoF wrist configuration. 
Since the arm was optimized to have as much of its inertia at its base, the weight of the \textit{new} 2-DoF wrist would be much heavier than its original configuration. In addition, its \textit{new} 3-DoF base would have small wrist actuators and consequently have lower maximum torque values that might not even be able to lift the arm itself. Effectively, a traditional robotic arm design would require favoring one end as the base or suffer a major reduction in its payload capacity. In the scenario where all actuators were of the same power and weight, one side would still be favored to maintain the fine orientation control at the wrist. Meanwhile, LIMMS has been designed to be as capable when anchored at either end.

\begin{figure}[h!]
\centering
 \begin{subfigure}[]{0.85\linewidth}
     \centering
     \includegraphics[width=\textwidth]{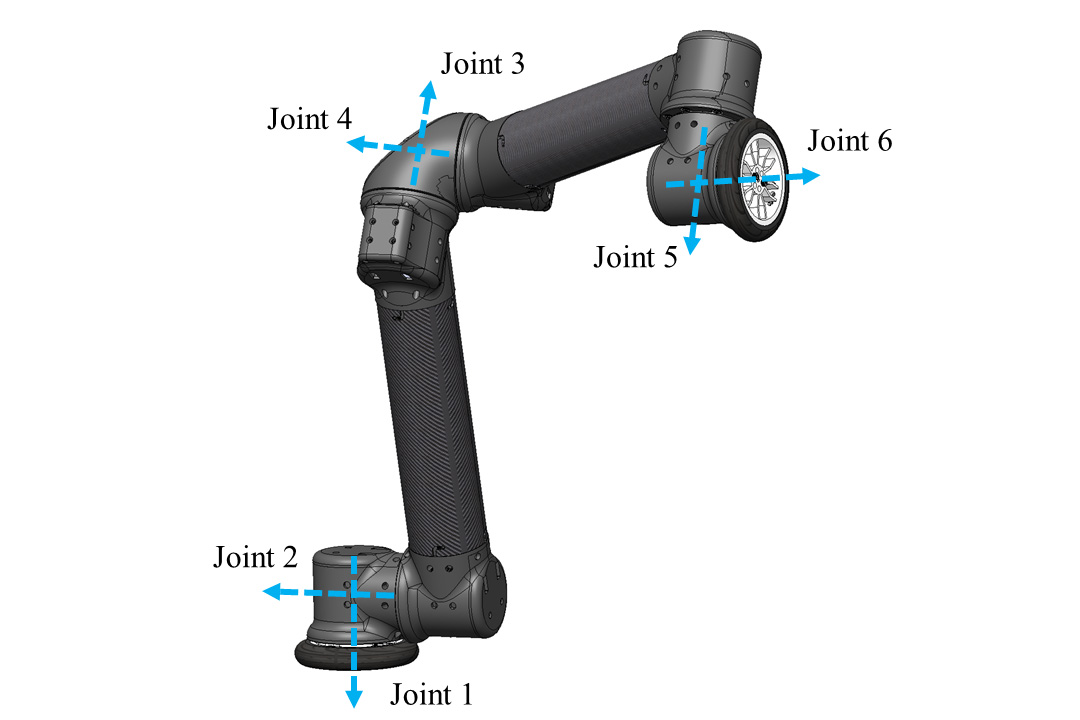}
     \caption{}
     \label{fig:joints}
 \end{subfigure}
 \hfill
 \begin{subfigure}[]{0.75\linewidth}
     \centering
     \includegraphics[width=\textwidth]{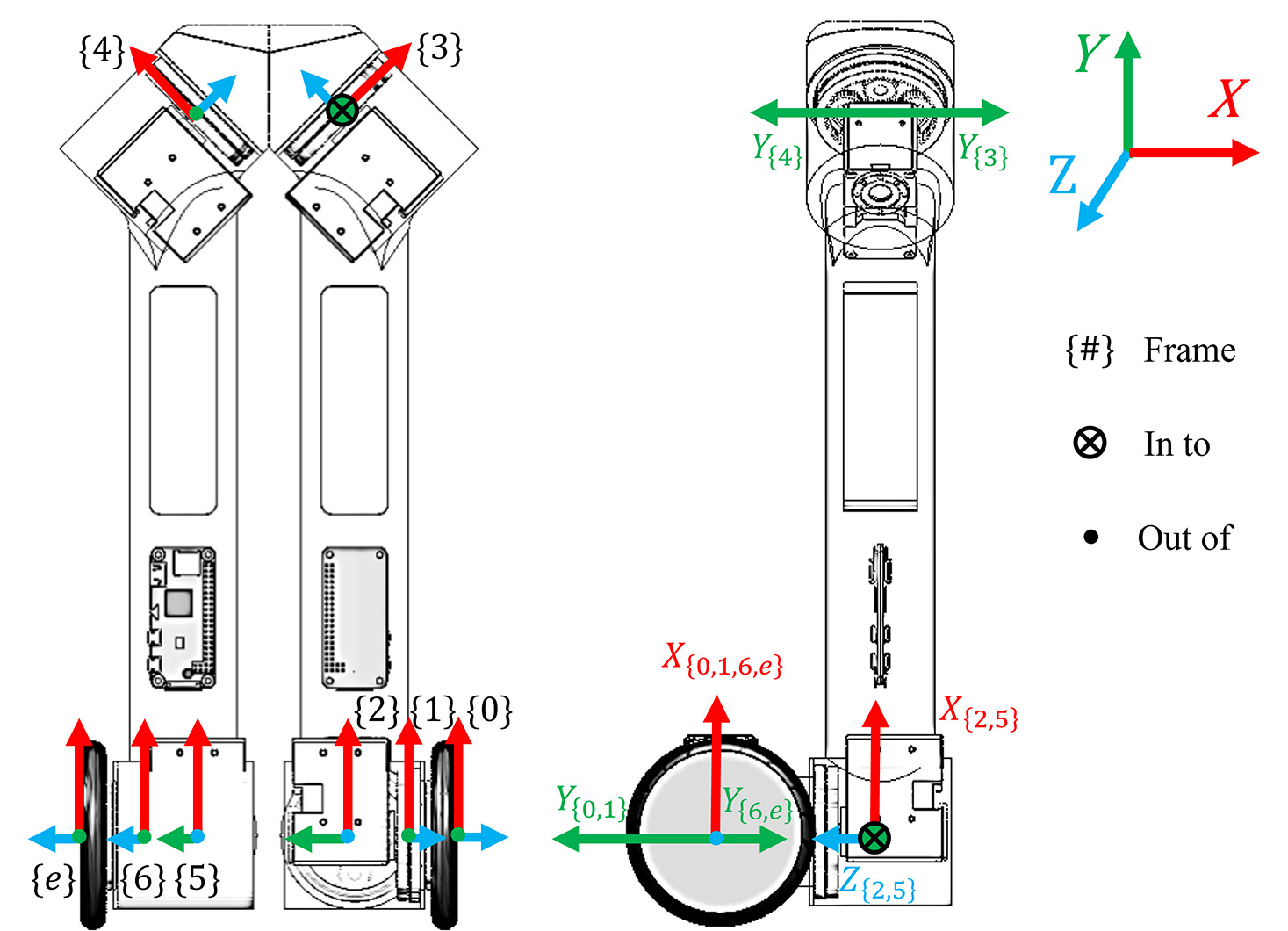}
     \caption{}
     \label{fig:kinematics}
 \end{subfigure}
 \caption{(a) LIMMS 6-DoF joint configuration. (b) \textit{Left} depicts a front view of LIMMS with joint frames. \textit{Right} shows a side view. Note that these are nontraditional DH frames.}
\label{fig:workspace_all}
\end{figure}

\begin{figure}[htb!]
\centering
 \begin{subfigure}[]{0.91\linewidth}
     \centering
     \includegraphics[width=\textwidth]{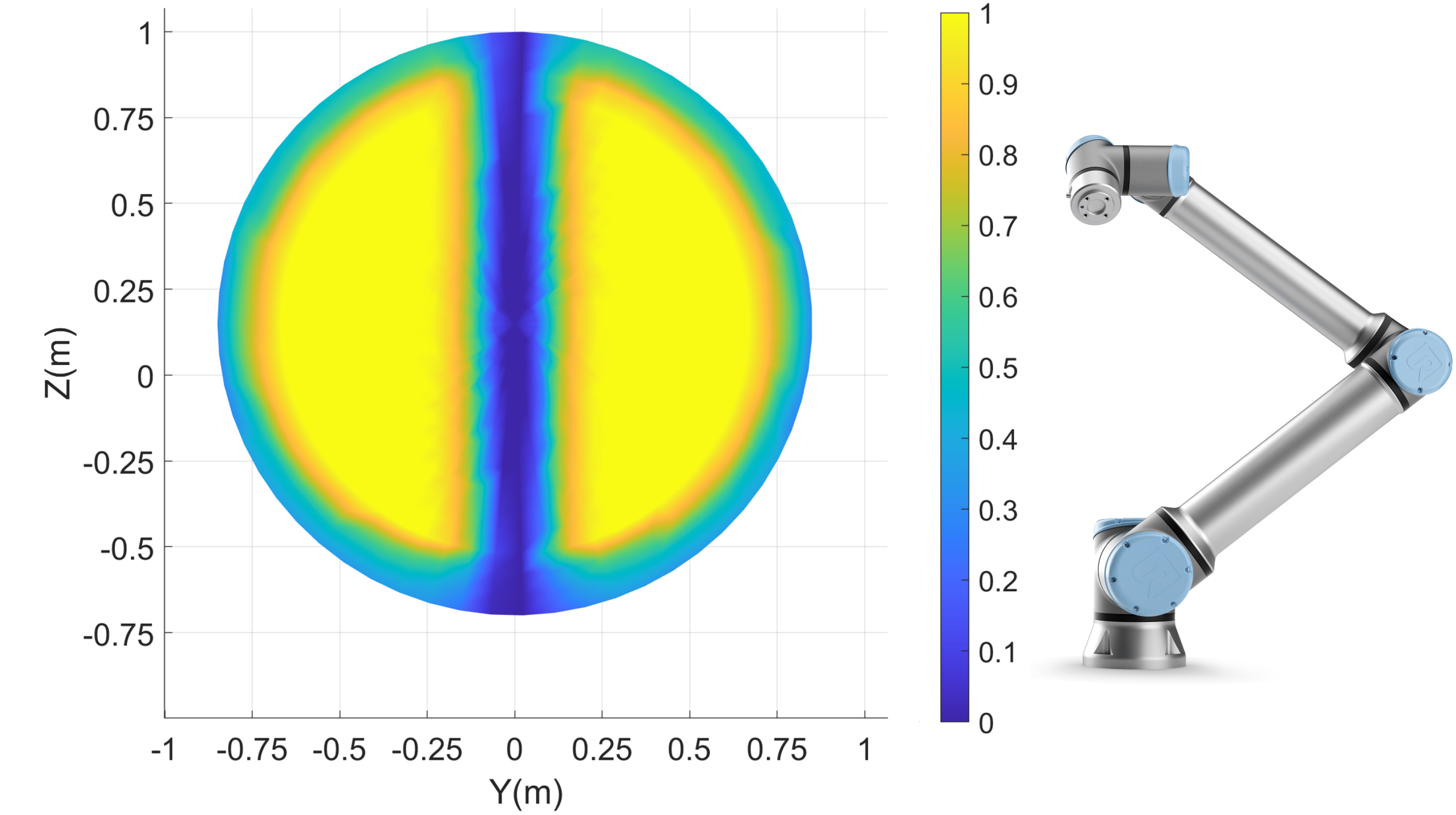}
     \caption{}
     \label{fig:dexterity_UR}
 \end{subfigure}
 \hfill
 \begin{subfigure}[]{0.91\linewidth}
     \centering
     \includegraphics[width=\textwidth]{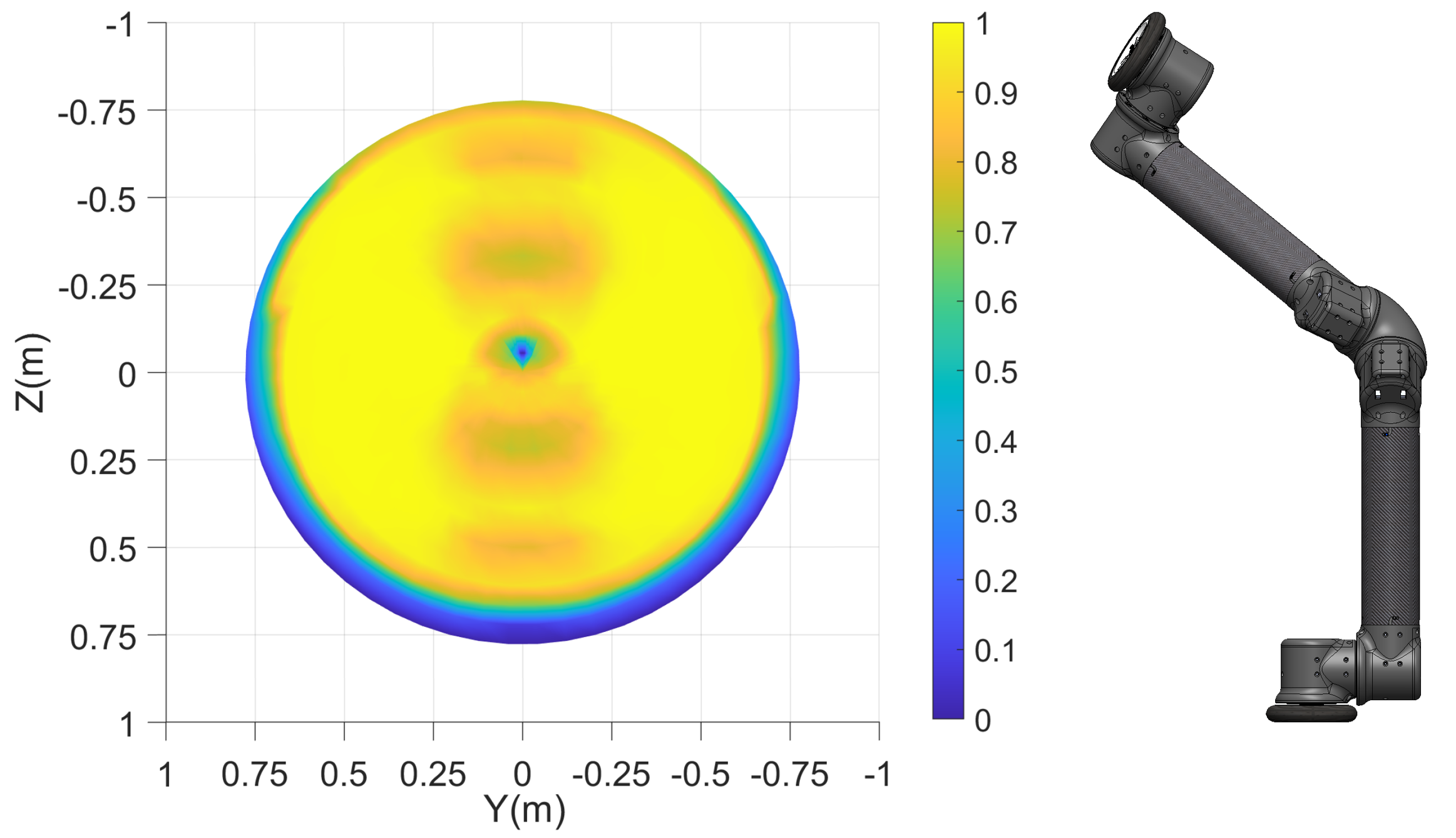}
     \caption{}
     \label{fig:dexterity_LIMMS}
 \end{subfigure}
 \caption{Dexterity index over the workspace \cite{dexterityindex} with a higher value, denoting the end effector can reach more orientations at a specified location, and a value of zero, implying the space cannot be reached, for: (a) Universal Robot UR5e \cite{UR}, a traditional 6-DoF robot manipulator, and (b) LIMMS, a nontraditional 6-DoF robot.}
\label{fig:dexterity_all}
\end{figure}

As a result, we designed LIMMS with a symmetric 6-DoF joint configuration as depicted in \cref{fig:joints}. LIMMS has 2 joints towards each end of the module and 2 joints near the elbow, leading to bilateral symmetry. 
This configuration can almost be viewed in a similar manner to a traditional 6-DoF arm where the elbow has to swing out to reach certain configurations. This serves as good intuition; however, it is a bit more involved. The $\frac{\pi}{4}$ offset from vertical at the elbow for joints 3 and 4 couples the position and orientation in a twisting fashion. Due to the pairing of these motors, the analytic inverse kinematics (IK) is nontrivial. Therefore, numerical IK was implemented using the Damped Least Squares method outlined in \cite{buss2005dlsik}. 

The frames laid out in \cref{fig:kinematics} do not directly follow the conventional Denavit–Hartenberg (DH) parameters. While it is possible to parametrize these frames using multiple intermediate rotations, the motivation for using non-standard frames was to preserve as much symmetry as possible and more naturally align the axes at the location of the motor output. More importantly, the frame locations and orientations do not change regardless of which end is used as a base. This establishes a more consistent intuition for the algorithm designer when switching between multiple anchor points in sequence. A side effect of the bilateral symmetric design is that the forward kinematics (FK) transformations are very similar in form, only requiring reversing the direction of a few offset angles when switching bases. For example, the transformation from joint 1 to joint 2 is the exact same calculation as the transformation from joint 6 to joint 5 except that the offset rotation around the X-axis is reversed. This results in a single FK algorithm with a boolean switch to solve for both configurations.

On the other hand, the unique configuration of LIMMS improves the dexterity of the workspace compared to other traditional 6-DOF robotic arm manipulators. \cref{fig:dexterity_UR} depicts the workspace dexterity for Universal Robot UR5e \cite{UR}. It is clear that the manipulator has low dexterity when the end effector approaches the outer edge of its workspace. Moreover, the region near the center axis has almost zero dexterity due to the offset between the two joints at the base. Even though LIMMS has a similar base configuration, the double joints at the elbow help improve the flexibility of the end-effector in the region near the center axis, as seen in \cref{fig:dexterity_LIMMS}.
 




\subsection{Latching}
\label{sec:latching}
Aligning and latching can quickly become a bottleneck in the entire delivery process, since the operation must be performed to switch between modes and manipulate objects. On top of that, real world factors, such as vibrations from the delivery vehicle engine and low lighting, could negatively influence alignment duration. While a camera and control algorithm are being investigated to coarsely align the end effector, the latching mechanism was designed to add mechanical robustness to misalignment.

\begin{figure}[th!]
\centering
 \begin{subfigure}[]{0.85\linewidth}
     \centering
     \includegraphics[width=\textwidth]{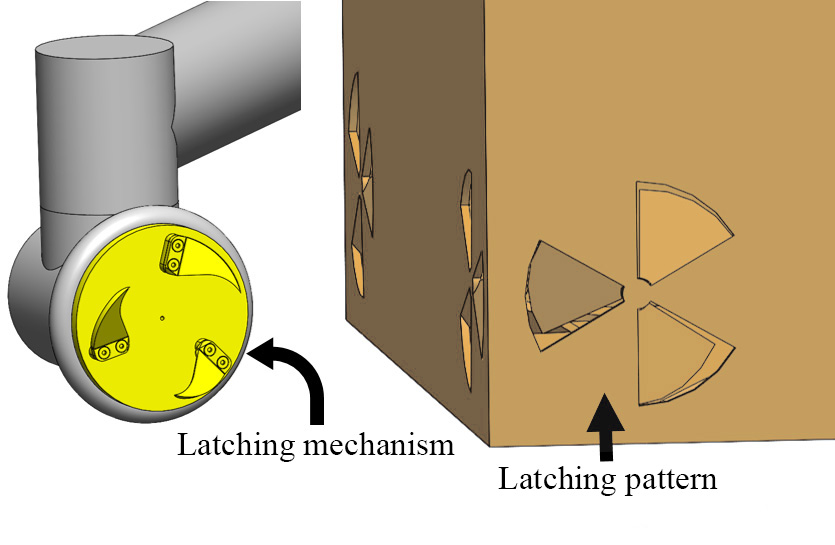}
     \caption{}
 \end{subfigure}
 \hfill
 \begin{subfigure}[]{0.7\linewidth}
     \centering
     \includegraphics[width=\textwidth]{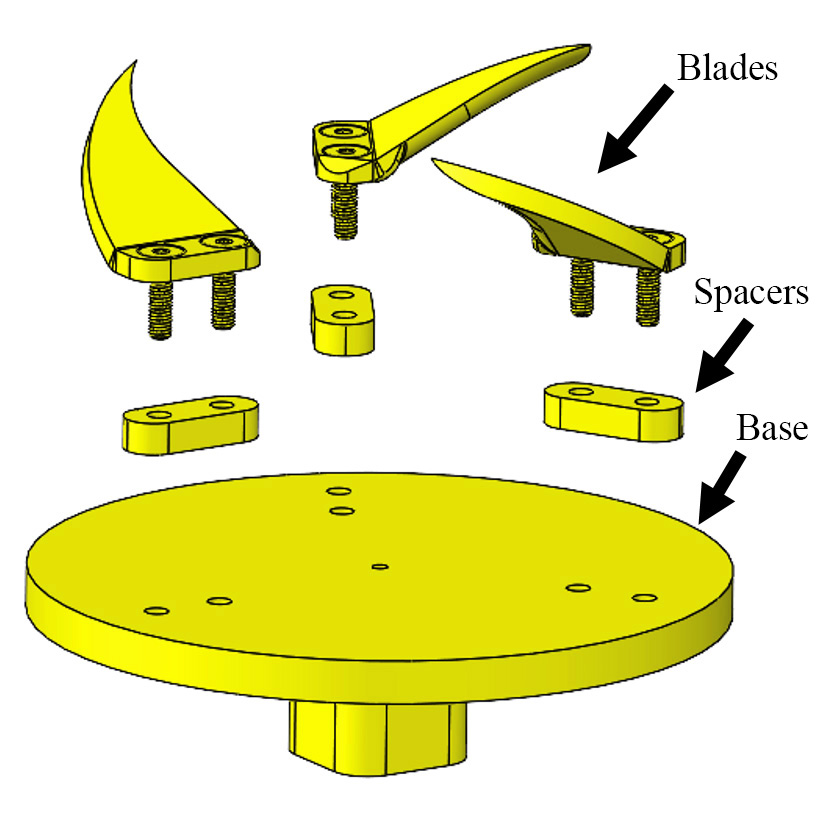}
     \caption{}
     \label{fig:latching_mechanism_blowout}
 \end{subfigure}
 \caption{(a) Latching mechanism that rotates into the mating hole pattern. (b) Breakout view of all of the latching prototype components.}
\label{fig:latching_mechanism}
\end{figure}

\begin{figure}[htb!]
\centering
 \begin{subfigure}[]{0.49\linewidth}
     \centering
     \includegraphics[width=\textwidth]{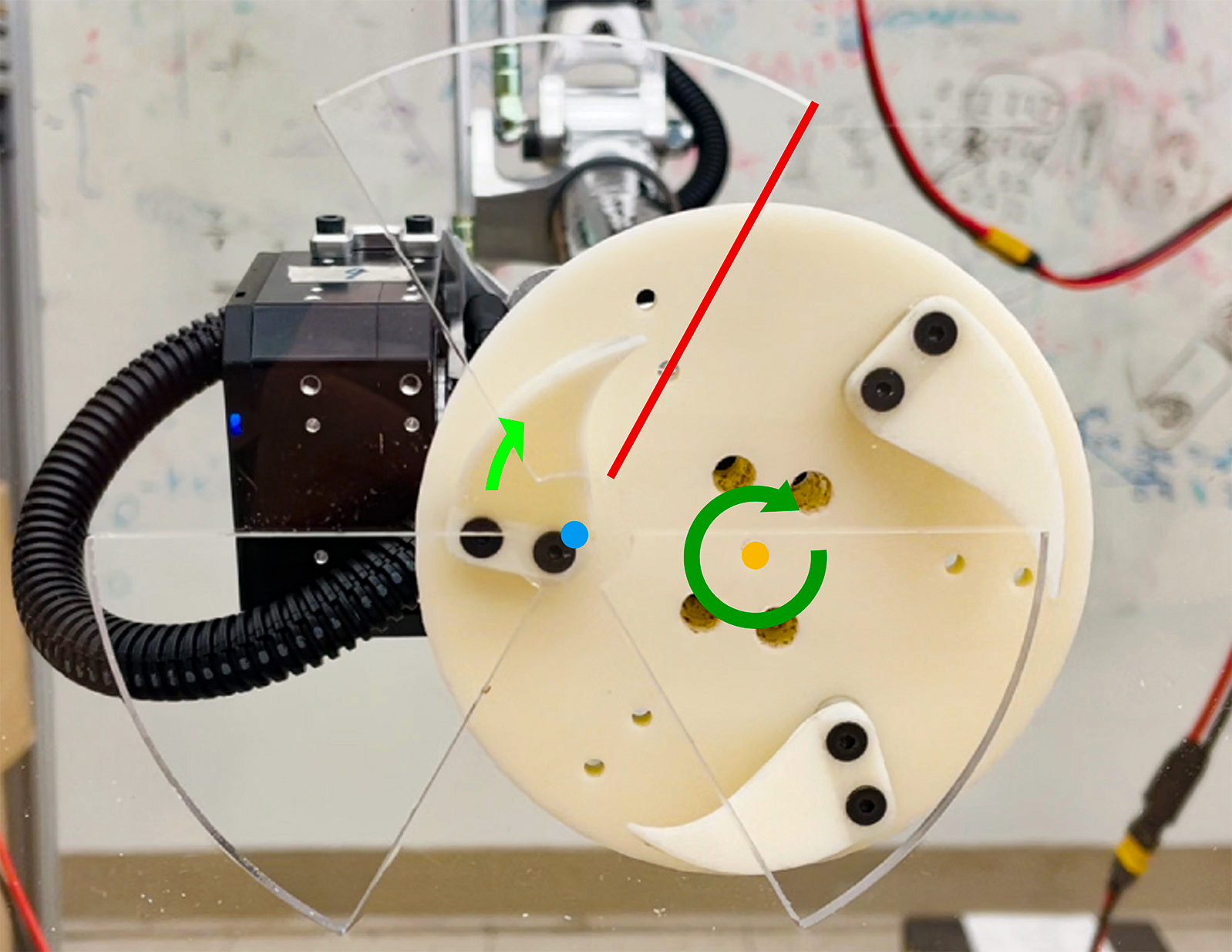}
     \caption{}
     \label{fig:latching_seq_zero_blade}
 \end{subfigure}
 \hfill
 \begin{subfigure}[]{0.49\linewidth}
     \centering
     \includegraphics[width=\textwidth]{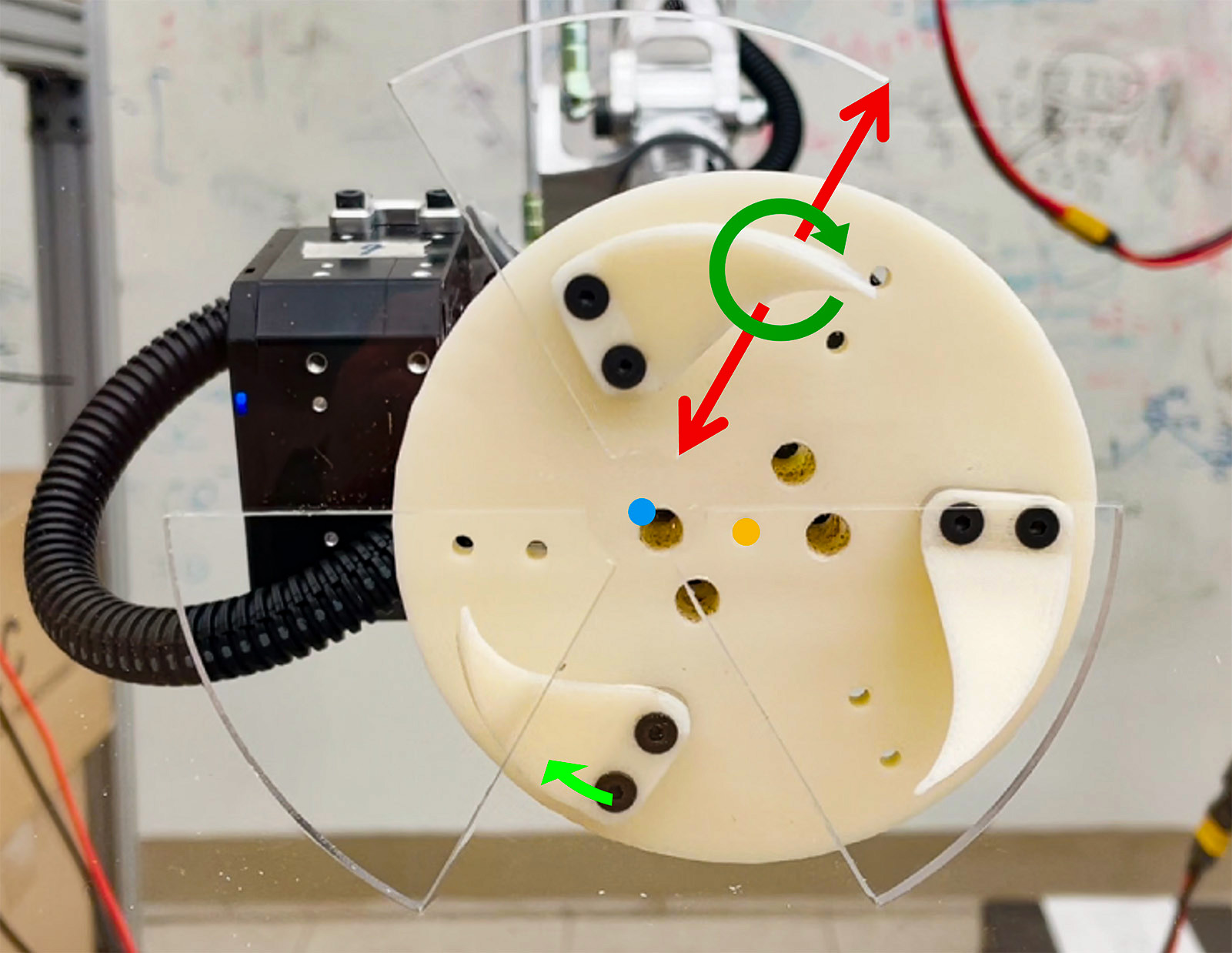}
     \caption{}
     \label{fig:latching_seq_one_blade}
 \end{subfigure}
 
 \begin{subfigure}[]{0.49\linewidth}
     \centering
     \includegraphics[width=\textwidth]{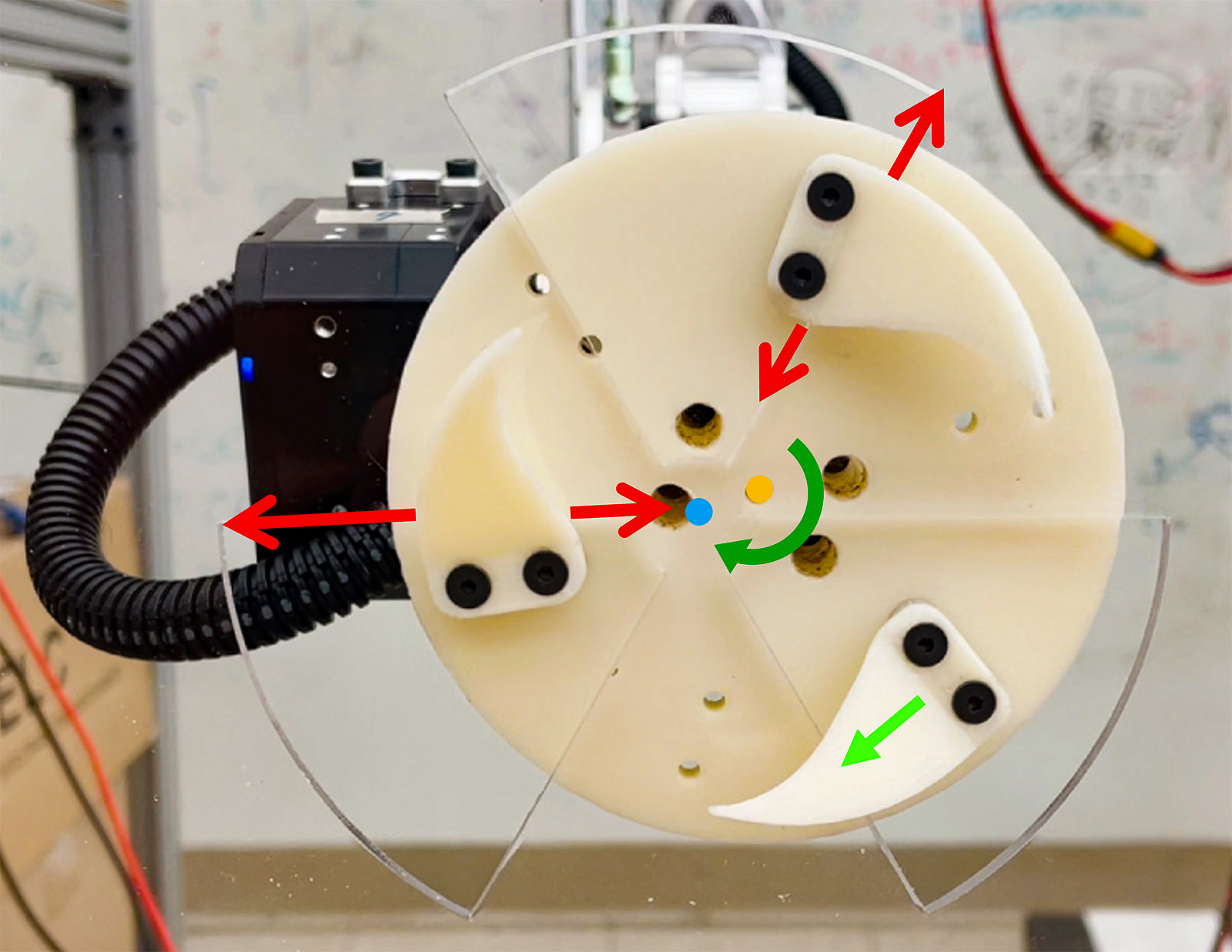}
     \caption{}
     \label{fig:latching_seq_two_blade}
 \end{subfigure}
 \hfill
 \begin{subfigure}[]{0.49\linewidth}
     \centering
     \includegraphics[width=\textwidth]{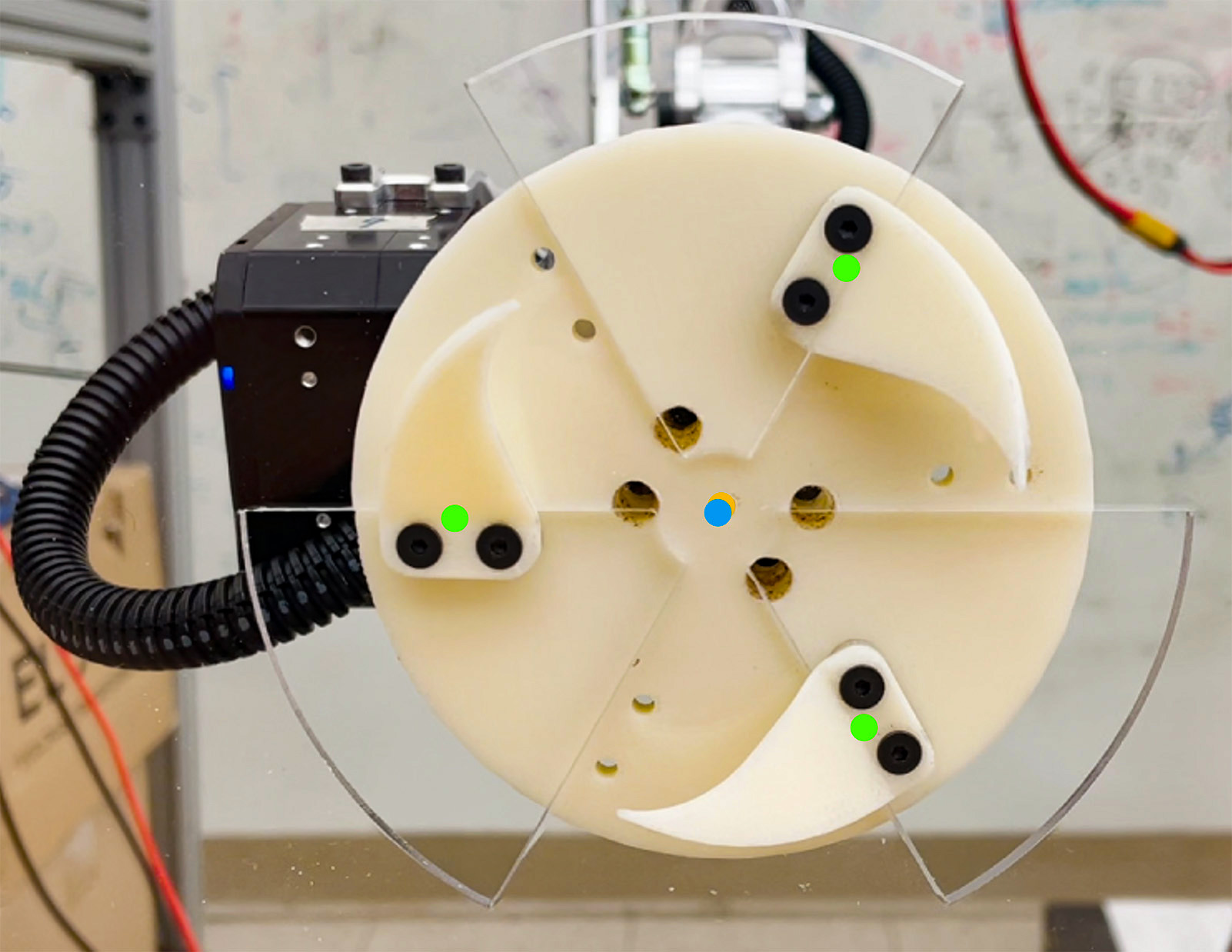}
     \caption{}
     \label{fig:latching_seq_three_blade}
 \end{subfigure}
\caption{LIMMS latching prototype self-aligning with initial misalignment into a sheet of acrylic with 3 cutouts: (a) Initial position with latch center (orange) rotating (dark green) until blade moving in direction (neon green arrow) until hits mating pattern edge (red) centered at (blue), (b) Contact blade becomes a point in which the mechanism can pivot about until the second blade engages, (c) 2 blades are constrained to moving along the red lines with arrows until the last blade makes contact, and (d) All blades are in contact (neon green dots) fully constraining the mechanism in an aligned position (orange and blue dots are concentric).}
\label{fig:latching_seq}
\end{figure}

Our initial prototype, as seen in \cref{fig:latching_mechanism}, has 3 blades that rotate into a cutout pattern of 3 triangular holes. The blades are sloped to pull the mechanism closer to the mating surface as the assembly is rotated. The spacers, as seen in the blowout view of \cref{fig:latching_mechanism_blowout}, are used to establish contacts with the edges of the triangular patterns to constrain the blade to linear movements along the edge face as the latching sequence progresses. The goal is that once all spacers have made contact with their respective mating edges, the assembly and the cutout patterns should be concentrically aligned and mechanically constrained. 

One possible alignment and mating sequence is shown in \cref{fig:latching_seq} where the mating surface is made of clear acrylic to better illustrate the process. The latching mechanism approaches the mating pattern in \cref{fig:latching_seq_zero_blade} but in gross misalignment as only one of the sloped blades successfully enters a triangular cutout. The entire assembly is rotated clockwise until the sloped blade makes contact with the edge of the hole (highlighted in red). Upon contact in \cref{fig:latching_seq_one_blade}, further rotation produces two movements: 1) clockwise rotation of the latching mechanism about the point of contact, bringing the second blade into alignment with the triangular cutout in the bottom left and 2) moving the mating surface along the slope of the blade, pulling the latching mechanism closer to the mating surface. 

Subsequent rotation causes the edge to contact with the spacer at the bottom of the blade as in \cref{fig:latching_seq_two_blade}. In this state, the edge of the triangular cutout and the flat face of the spacer are flush and held together by the clockwise torque of the assembly. Due to the angle between the triangular patterns, the latching assembly is constrained to travel along the contact edge, moving in the upper right hand direction. Meanwhile, the second blade has now moved into the second triangular cutout, pulled the mating surface in, and the spacer has made contact with the second triangular edge. 

With these two linear constraints, the final clockwise rotation will align the assembly, causing the last blade and the last spacer to make contact and fix the mating pattern to the assembly face as in \cref{fig:latching_seq_three_blade}. With this the alignment process is complete and LIMMS is attached to the mating surface. It is important to note that while the above explanation was broken into steps, the entire procedure with the hardware occurred at a continuous speed without stopping. 

Another important distinction is that with this prototype, the assembly must still apply torque to maintain the contact. The main goal of the prototype was to verify the effectiveness of the alignment procedure. To prevent radial movement and lock the latch in place, future iterations will have linear pins that extend from the latching mechanism through the mating surface. Thus, LIMMS will be free to use the end actuator to rotate while maintaining the fixture to the surface. 

Through this design, LIMMS can tolerate a high degree of position and orientation misalignment in both the the initial pose of the latch and its mating surface. Multiple design considerations were made to afford this level of robustness. By increasing the cutout area relative to blade size, the first blade has a much higher probability of finding any given hole to start the process. Meanwhile, the slope and shape of the blades can produce some self-aligning alone and while not strictly necessary, still provides the system with more tolerance for error. Concurrent work is being done to further optimize these designs to increase the latching mechanisms robustness and details the process in more depth \cite{latching2022}.

In addition to the experiment in \cref{fig:latching_seq}, testing with cardboard mating surfaces was conducted and was successful for intially misaligned configurations. A key design feature of the 3 blade design is to place loads only on the radially outward edges of the hole pattern. This prevents the center \textit{island} connecting all three holes on the mating surface from experiencing direct loads and causing damage to the pattern. Based on the payload, the center \textit{island} can be increased in size to ensure structural stability. For the purposes in the experiment, an 8 mm radius was sufficient.

\begin{figure}[t]
\centering
 \begin{subfigure}[]{1\linewidth}
     \centering
     \includegraphics[width=\textwidth]{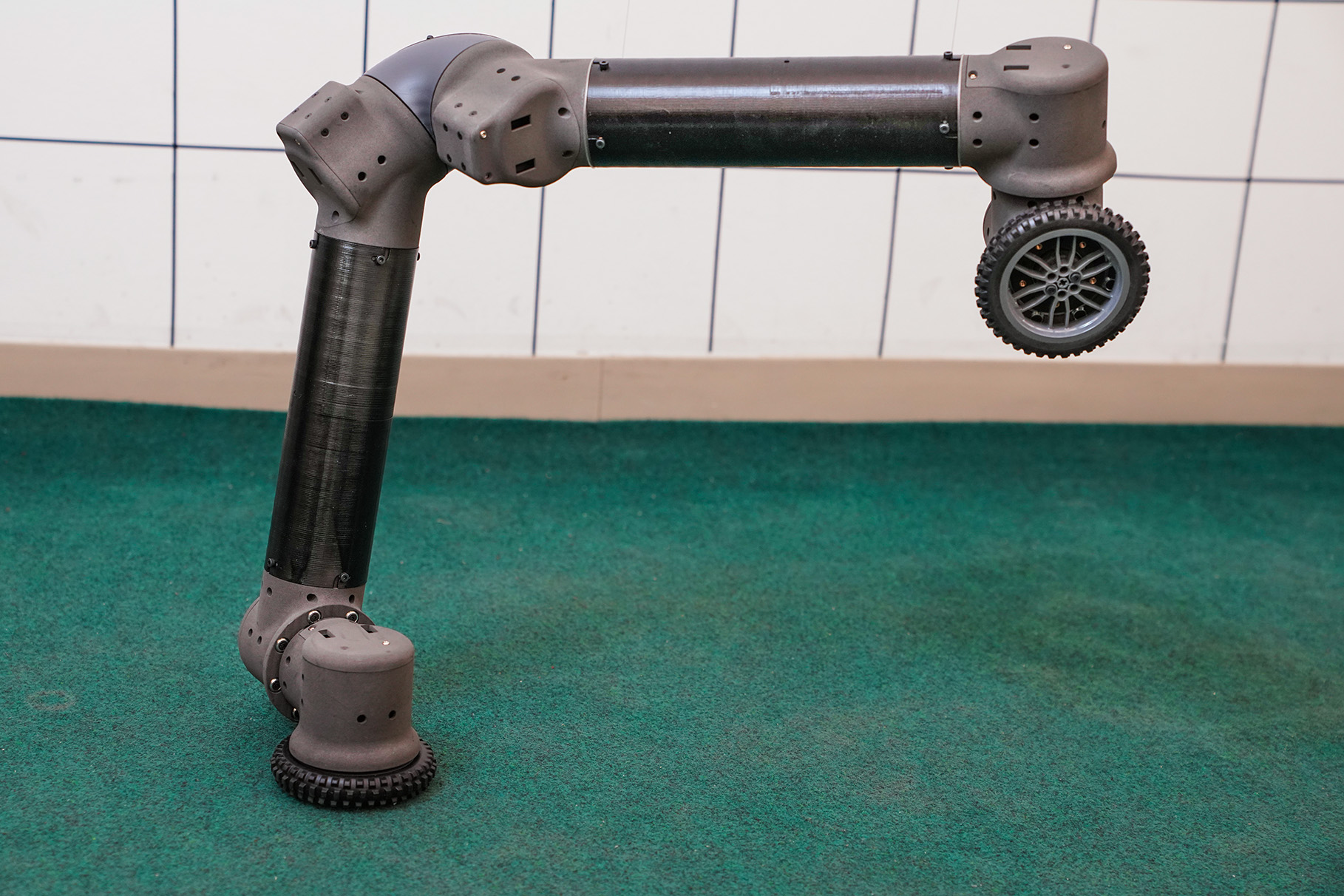}
     \caption{}
     \label{fig:prototype}
 \end{subfigure}
 \hfill
 \begin{subfigure}[]{1\linewidth}
     \centering
     \includegraphics[width=\textwidth]{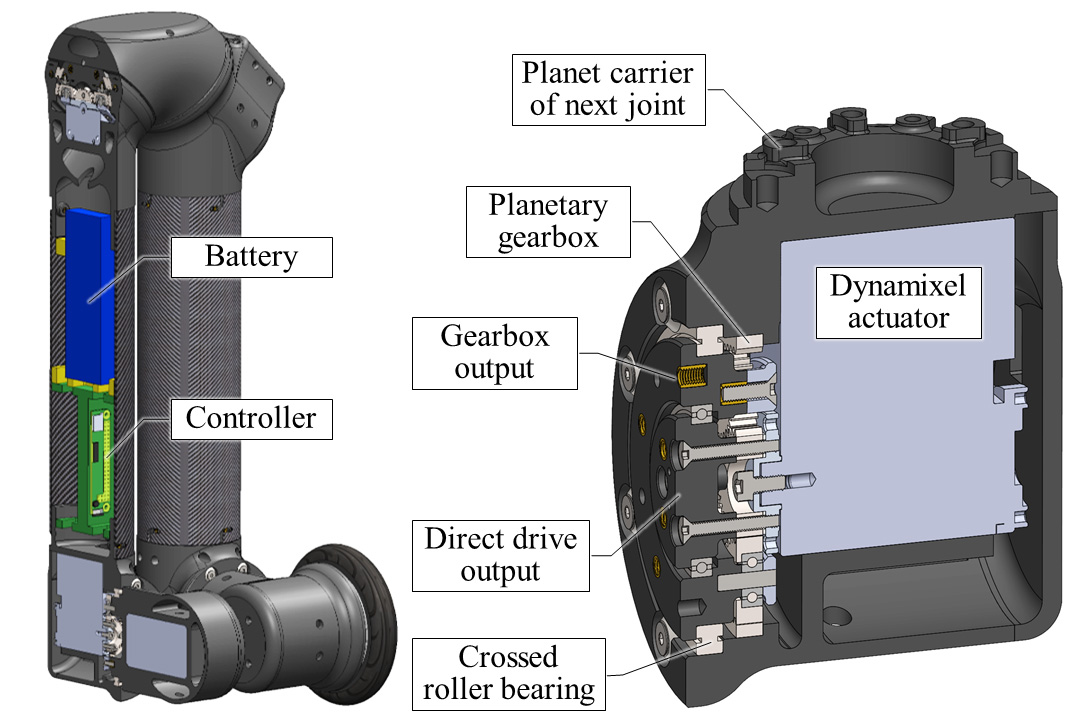}
     \caption{}
     \label{fig:prototype_cad}
 \end{subfigure}
 \caption{(a) LIMMS hardware prototype. (b) Section view of LIMMS prototype using commercial off-the-shelf actuator with custom gearbox.}
\label{fig:hardware_all_figs}
\end{figure}

\subsection{Hardware Prototype}
\label{sec:prototype}

\begin{figure*}[h!]
    \centering
     \begin{subfigure}[]{0.32\linewidth}
         \centering
         \includegraphics[width=\textwidth]{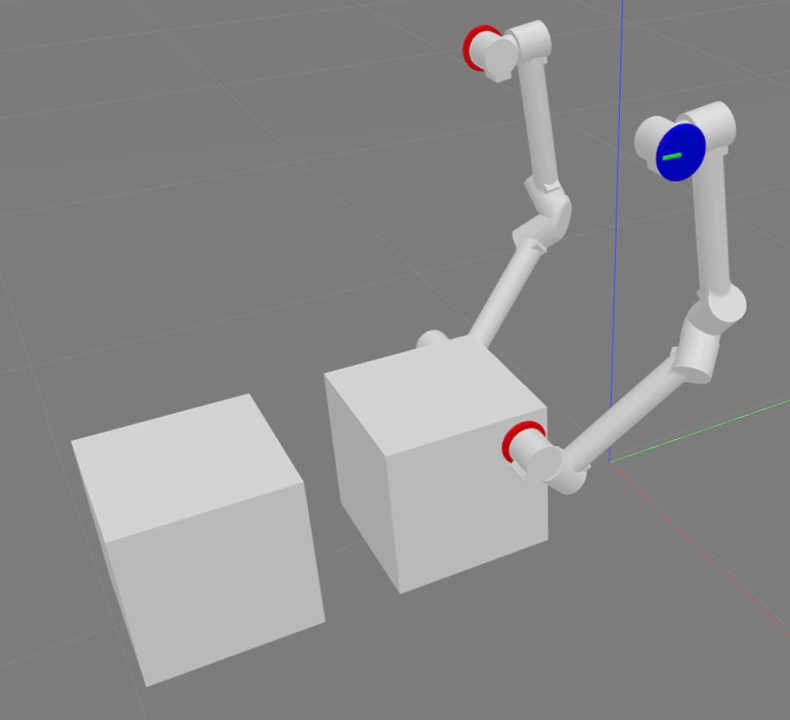}
         \caption{$t=0s$}
         \label{fig:manipulation_start}
     \end{subfigure}
     \begin{subfigure}[]{0.32\linewidth}
         \centering
         \includegraphics[width=\textwidth]{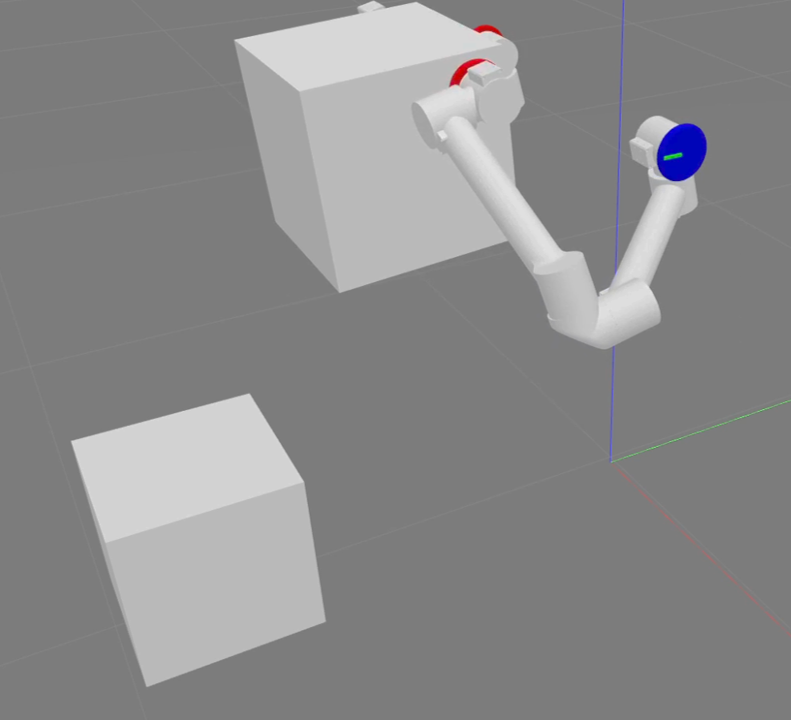}
         \caption{$t=5s$}
         \label{fig:manipulation_middle}
     \end{subfigure}
     \begin{subfigure}[]{0.32\linewidth}
         \centering
         \includegraphics[width=\textwidth]{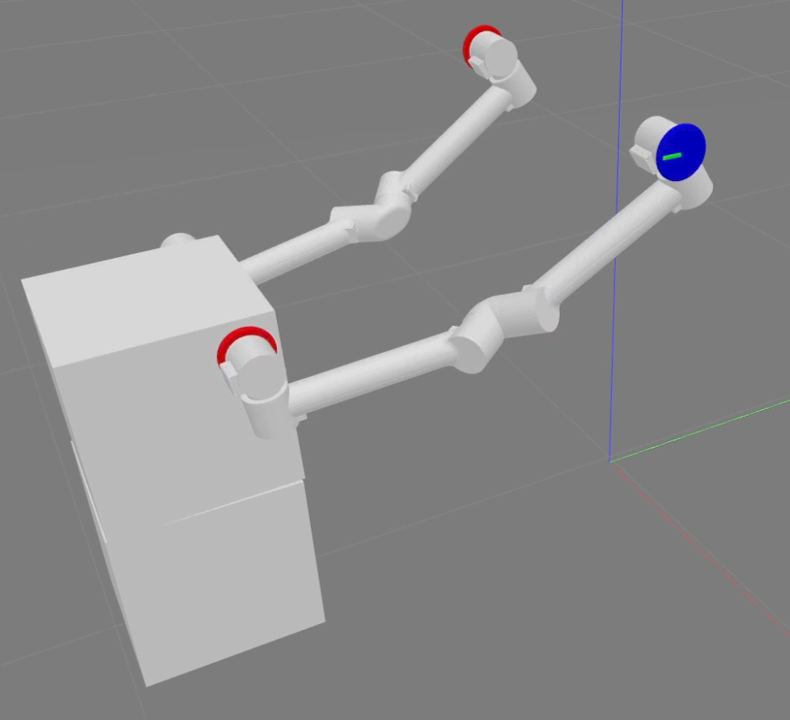}
         \caption{$t=10s$}
         \label{fig:manipulation_end}
     \end{subfigure}
     \begin{subfigure}[]{0.48\linewidth}
         \centering
         \includegraphics[width=\textwidth]{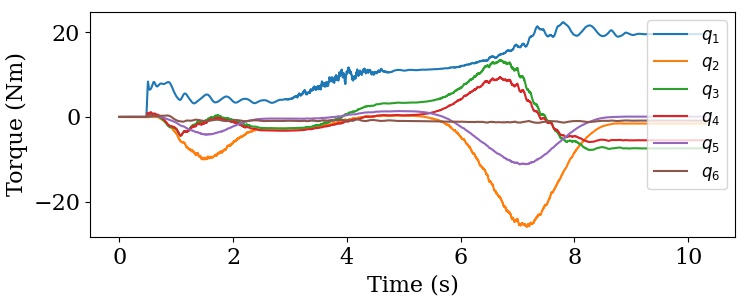}
         \caption{Torque of the joints on a single LIMMS.}
         \label{fig:manipulation_torque}
     \end{subfigure}
     \hfill
     \begin{subfigure}[]{0.48\linewidth}
         \centering
         \includegraphics[width=\textwidth]{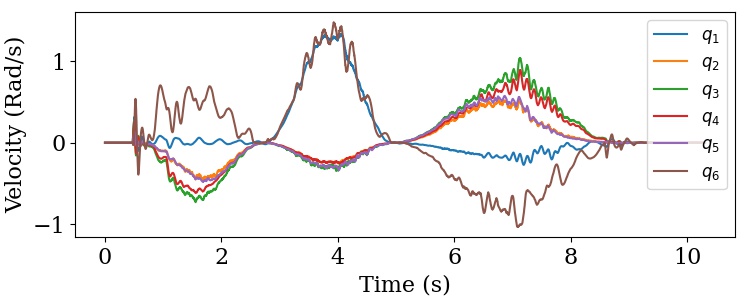}
         \caption{Velocity of the joints on a single LIMMS.}
         \label{fig:manipulation_velocity}
     \end{subfigure}
    \caption{Two LIMMS where the bases are assumed to be anchored in the vertical plane lifting a box.}
    \label{fig:manipulation}
 \end{figure*}

The first hardware prototype of LIMMS was designed and manufactured, as shown in \cref{fig:prototype}, to evaluate its unique kinematics design. The prototype is constructed using commercial off-the-shelf actuators and rapid prototyping methods. Carbon fiber tubes house the battery and the controller while the structural parts of the prototype joints have been manufactured from nylon 12 using Selective Laser Sintering (Fuse 1 SLS printer, Formlabs). This facilitates quick investigation of different link lengths and sizes for future prototypes.


Each joint is actuated by a DYNAMIXEL XM540-W150-R motor from ROBOTIS.
Due to the specifications of these actuators, extra external speed reduction was needed to reach the target torque. As shown in \cref{fig:prototype_cad}, a planetary gearbox with 3.5:1 reduction ratio has been implemented into the design to achieve 31 Nm of peak torque with a 2 rad/s max velocity.
Additionally, crossed roller bearings (CRBT505A, IKO) were used for the output planet carrier to maximize rigidity.

For the hardware prototype, either the wheels or latching mechanism can be mounted onto the end joints. By designing a swappable end effector, multiple configurations can be fully evaluated before finalizing the design. The latching mechanism is mounted to the output carrier of the planetary gearbox, while the wheel is attached directly to the actuator. This gives wheels higher velocity for locomotion (88517 \& 11957, LEGO). 



\section{Task simulation}
\label{sec:simulation}
Feasibility studies of delivery sub-tasks were performed in simulation to investigate the capability of the LIMMS modules as well as verify hardware choices made for the prototype. All simulations used the open source Gazebo simulator \cite{gazebo}. To simulate parts of the delivery sequence, the sub-tasks chosen were as follows:
\begin{enumerate}
    \item Two LIMMS as dual manipulators,
    \item Four LIMMS attached to a box as a quadruped, and
    \item Single LIMMS in self-balancing wheeled mode.
\end{enumerate}
For each of these modes, the torque and velocity at each joint were analyzed over a given trajectory. 
It is important to note that LIMMS is not only restricted to these operating modes, as discussed in \cref{sec:conclusion}. Rather, these sub-tasks were chosen to substitute as a minimal threshold for completing a delivery and explore two of the more demanding tasks.

\begin{figure*}[h!]
    \centering
     \begin{subfigure}[]{0.245\linewidth}
         \centering
         \includegraphics[width=\textwidth]{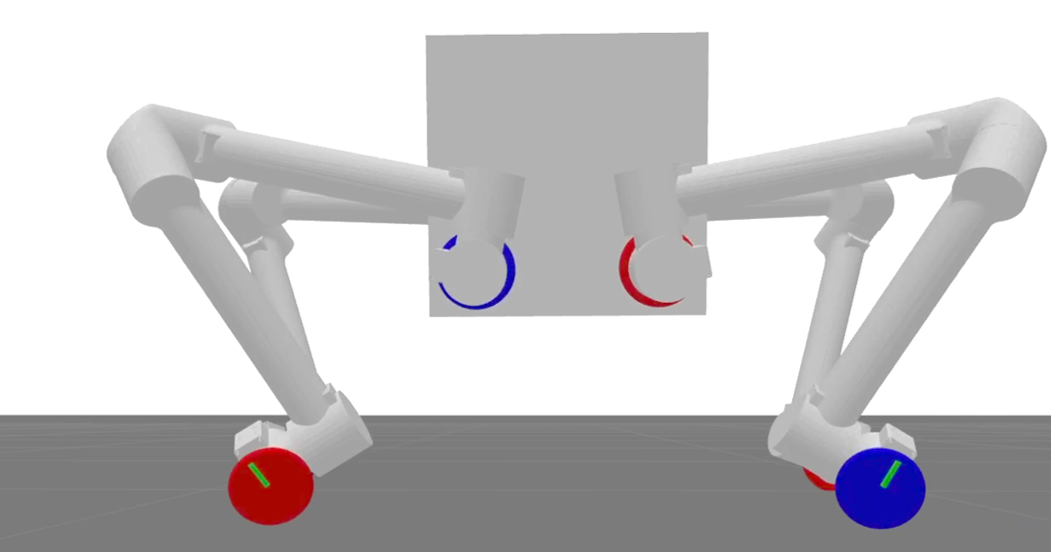}
         \caption{$t=0s$}
         \label{fig:walk0}
     \end{subfigure}
     \hfill
     \begin{subfigure}[]{0.245\linewidth}
         \centering
         \includegraphics[width=\textwidth]{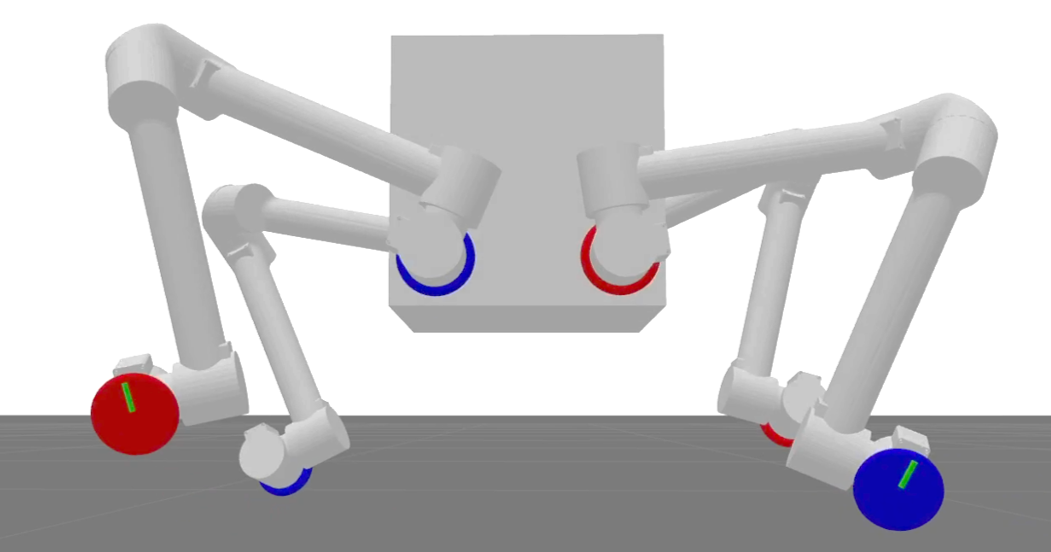}
         \caption{$t=0.25s$}
         \label{fig:walk1}
     \end{subfigure}
     \hfill
     \begin{subfigure}[]{0.245\linewidth}
         \centering
         \includegraphics[width=\textwidth]{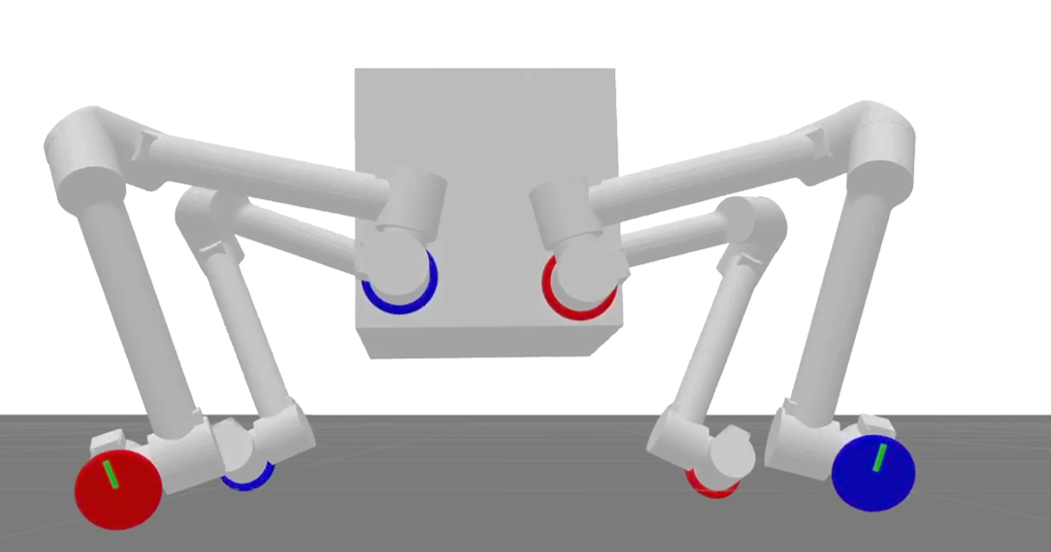}
         \caption{$t=0.5s$}
         \label{fig:walk2}
     \end{subfigure}
     \hfill
     \begin{subfigure}[]{0.245\linewidth}
         \centering
         \includegraphics[width=\textwidth]{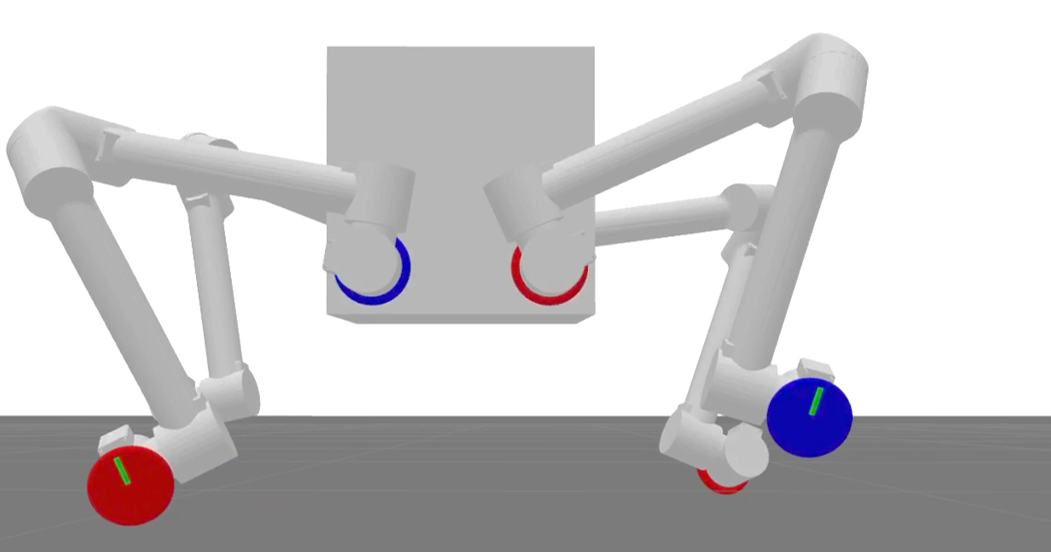}
         \caption{$t=0.75s$}
         \label{fig:walk3}
     \end{subfigure}
     \hfill
     \begin{subfigure}[]{0.48\linewidth}
         \centering
         \includegraphics[width=\textwidth]{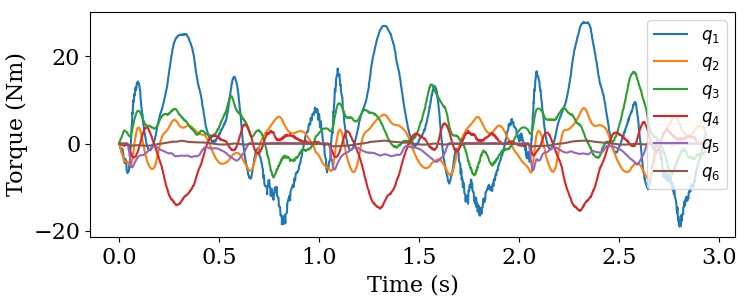}
         \caption{Torque of joints on a single LIMMS.}
         \label{fig:quadruped_torque}
     \end{subfigure}
     \hfill
     \begin{subfigure}[]{0.48\linewidth}
         \centering
         \includegraphics[width=\textwidth]{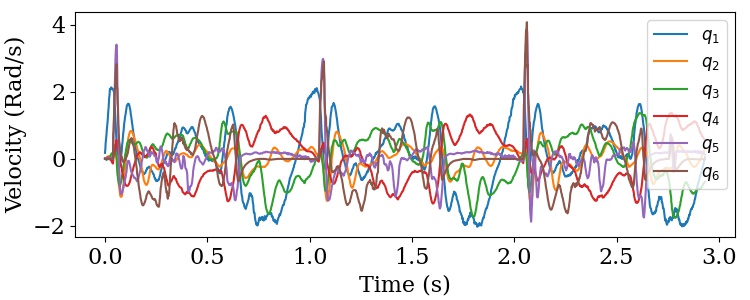}
         \caption{Velocity of joints on a single LIMMS.}
         \label{fig:quadruped_velocity}
     \end{subfigure}
     \hfill
    \caption{Four LIMMS attached to a 2kg box performing a trot gait. Frames of the trot gait were taken starting from (a) through (e).}
    \label{fig:sim_quadruped}
 \end{figure*}

As described in \cref{sec:introduction}, LIMMS would mount to mating patterns within the delivery vehicle to perform manipulation tasks such as taking boxes off of a stack, moving boxes towards the exit, or restacking boxes to keep the truck organized. To verify this, two LIMMS were mirrored with each other and anchored midair to simulate a simple lifting motion of a 2 kg box as seen in \cref{fig:manipulation}. Each LIMMS actuator was controlled in a decentralized manner with a simple position PID (proportional–integral–derivative) controller outputting torque commands at each joint. The lift up and stack motion was performed using three key frames in operational space. Joint position commands were then sent using joint space interpolation between key frames.

The corresponding torque and velocity graphs for a single LIMMS module during the motion can be seen in \cref{fig:manipulation_torque} and \cref{fig:manipulation_velocity}, respectively. The mirrored LIMMS module experienced a similar load. As expected, the base motor requires the most torque, especially as the 2kg box is moved further away creating a large torque arm. The large magnitude of joint 2 is of interest as there should not be any large forces acting perpendicular to the box movement. This suggests the LIMMS module is pulling away from the box and might be due to lack of coordination between the two modules from the simple control scheme used. The max torque required was around 25 Nm which was just below the hardware specification. More complex control algorithms for coordination and trajectory planners which optimize the choice of anchor points to perform manipulation closer to the base will be investigated to lower peak torque requirements.


 
Meanwhile, four LIMMS, each attached to a different corner of a box, were simulated to walk as a single unit as depicted in \cref{fig:sim_quadruped}. The overall system performed a trot gait using Raibert heuristics for footstep planning and Bezier interpolated swing leg trajectories as described in \cite{hooks2020alphred}. The trot gait was done with a stance period of 0.5 s and swing period of 0.5 s. From the gait planning, the joint positions were fed into the decentralized position PID controllers. While a stance phase was not planned, the LIMMS quadruped can be seen transitioning from stance (\cref{fig:walk0}), to lifting both red feet up (\cref{fig:walk1}), back to stance  (\cref{fig:walk2}), and finally lifting both blue feet up (\cref{fig:walk3}). This gait cycle was repeated indefinitely to move at 0.3 m/s. This was all accomplished using open loop trajectories without consideration for leg dynamics.


 
Three gait cycles worth of torque and velocity data from the quadruped trot experiment can be seen in \cref{fig:quadruped_torque} and \cref{fig:quadruped_velocity}, respectively. The torque maxed out at 28.6 Nm and was near the limit of the hardware. Meanwhile, the velocity peaked at 4.1 m/s, but only for 0.005s which suggests that trotting on the hardware requires smoother control. The base motor again experiences the most load as it needs to lift the entire leg during swing phase. Unlike most conventional quadrupeds where the leg inertia is minimized \cite{cheetah3}, LIMMS has its inertia spread out through the module. Since the leg dynamics were not considered, this made following the swing leg trajectory more difficult particularly at the beginning and ends of the sequence when lifting off or touching down. As a result, the quadruped system experienced an unplanned stance phase that is otherwise unseen in regular trot gaits. In addition, a single LIMMS module weighs roughly twice as much as the body. As most quadruped control systems assume massless legs \cite{hooks2020alphred}, a new control schema may be necessary to account for the atypical weight distribution. This problem was most evident on each touchdown event as the entire body would bounce from the momentum exchange. Consequently, the gait was sensitive to variations in gait timing, step height, and desired velocity. Further work must be done to incorporate leg dynamics and more complex coordination between to the legs to achieve more robust performance.



The final sub-task simulated was locomotion through a self-balancing wheeled mode. By commanding all actuators other than the wheels to hold zero positions, LIMMS can operate in a similar manner to an underactuated two wheel mobile robot, eg., a Segway \cite{segway}. The desired wheel speed velocity can be maintained while balancing the body using PD controllers \cite{gao2012dynamics}. This provides a single LIMMS with a simple locomotion strategy without anchors or other LIMMS.

From the first two simulations, it is evident that proper coordination between LIMMS modules must be addressed. Adding a LIMMS to a task should increase the potential lifting capacity of the combined system. However, without an efficient cooperation strategy, this can have an adverse impact where the LIMMS modules end up disrupting task completion. Developing an algorithm for this is considered to be outside the scope of this paper but will developed in future work. Overall, the three simulated sub-tasks verified and demonstrated the feasibility while providing more intuition into the shortcomings that need to be addressed in the future.
\section{Conclusion}
\label{sec:conclusion}

In this paper, LIMMS was introduced as a modular, multi-modal robotics concept for last-mile delivery. Based on real-world data, the design requirements and assumptions were laid out and tested. The LIMMS kinematics and design were then shown to have a better workspace than a traditional 6-DoF robotic arm and not suffer from asymmetric torque imbalances. A rotational self-aligning mechanism was also introduced to be used at the end effectors of LIMMS. A proof-of-concept hardware prototype for both a LIMMS unit and a latch were built. Finally, sub-tasks of the delivery process were carried out in simulation for the feasibility study.



LIMMS fills a niche between the general purpose and task-specific robots. While it is suited mainly for manipulation tasks, it still remains adaptable for application on most other delivery tasks. The compact design allows multiple LIMMS to replace larger, complex systems. This comes at a cost. For instance, the wheeled delivery robot may be more energy efficient than the LIMMS quadruped unit. However, the wheeled robot is effectively constrained to a specific type of terrain. LIMMS is not. For a humanoid delivery robot, the arms and legs of the robot are constrained by the physical structure of its back and hip widths. This may be a disadvantage for very heavy or awkwardly shaped payloads as those spacing constraints limit the methods in which to carry the box and how many humanoids work together on a single box. LIMMS can potentially scale up to carry heavier loads than a humanoid because it does not have the same physical spacing constraints. Moreover, when scaling the number of LIMMS units it can squeeze into a number of places and attach itself on walls or even roofs. 


The generalizability of LIMMS additionally opens up new avenues of optimization for logistics never before seen. As long as there are attachment points, LIMMS can latch its base link to it, meaning that for warehouses there is a potential to not only optimize traffic of automated systems on the ground but vertically as well. Mixing and matching all of the modes can bring about much more efficient delivery logistics. For example, the wheels on LIMMS could be combined with manipulator mode to form a dynamic mobile conveyor belt system to move stacks of boxes.

The LIMMS prototype presented in this paper only provides us with a test bed to help us better understand our design choices, torque and velocity requirements, multi-modal controllers, and latching capabilities. As such, some features are missing, such as a simultaneous wheel and latch actuation and a locking mechanism for the latch. We plan on running more preliminary tests and then updating the design accordingly while implementing these new features for a new prototype. In the process we aim to develop better control algorithms and high-level planners to efficiently deliver packages in the last mile.





\section*{Acknowledgements}
The authors would like to thank LG Electronics for sponsoring this research and providing useful feedback.

{
\bibliographystyle{IEEEtran}
\bibliography{references}
}


\end{document}